\documentclass[twoside,11pt]{article}

\usepackage[preprint]{jmlr2e}
\usepackage{amsmath}
\usepackage{algorithm}
\usepackage{algorithmic}

\def\M{{\cal M}}
\def\S{{\cal S}}

\def\T{{\cal T}}
\def\V{{\cal V}}
\def\R{{\mathbb{R}}}

\firstpageno{1}

\title{Semi-Supervised Anomaly Detection Based on \\
Quadratic Multiform Separation}

\author{\name Ko-Hui Michael Fan\thanks{Corresponding author}
\email mkhfan@gmail.com\\
\addr AEM Technology Taiwan Corporation\\
Taipei, Taiwan, R.O.C.
\AND
\name Chih-Chung Chang \email ccchang@math.ntu.edu.tw\\
\addr Department of Mathematics\\
National Taiwan University\\
Taipei, Taiwan, R.O.C.
\AND
\name Kuang-Hsiao-Yin Kongguoluo \\
\addr AEM Technology Taiwan Corporation\\
Taipei, Taiwan, R.O.C.
}

\begin{document}
\maketitle
\begin{abstract}
In this paper we propose a novel method 
for semi-supervised anomaly detection (SSAD).
Our classifier is named QMS22 as its inception was dated 2022 upon the
framework of quadratic multiform separation (QMS), a recently introduced
classification model. QMS22 tackles SSAD
by solving a multi-class classification problem involving
both the training set and the test set of the original problem.
The classification problem intentionally includes
classes with overlapping samples. One of the classes contains
mixture of normal samples and outliers, and all other classes
contain only normal samples. An outlier score is then calculated
for every sample in the test set using the outcome of the
classification problem. We also include performance evaluation
of QMS22 against top performing classifiers using ninety-five benchmark
imbalanced datasets from the KEEL repository. These classifiers
are BRM (Bagging-Random Miner), OCKRA (One-Class K-means with
Randomly-projected features Algorithm), ISOF (Isolation Forest),
and ocSVM (One-Class Support Vector Machine). It is shown by
using the area under the curve of the receiver operating
characteristic curve as the performance measure,
QMS22 significantly outperforms ISOF and ocSVM. Moreover, the
Wilcoxon signed-rank tests reveal that there is no statistically
significant difference when testing QMS22 against BRM nor
QMS22 against OCKRA.
\end{abstract}

\section{Introduction}
Anomaly detection is the identification of rare items or outliers
in a dataset. As a learning task, it
may be supervised, semi-supervised, or unsupervised.
Anomaly detection is an important data mining task that has been
studied within diverse research areas.
Inherently, outliers represent some kind of problem
such as network intrusion, bank fraud, faulty equipment, or heart arrhythmia.
Moreover, anomaly detection is often used in preprocessing to remove
outliers from the dataset, which benefits
the machine learning tasks that follow.
There is a large number of survey papers on anomaly detection in the literature
and we refer the readers to
\citep{10.1145/1541880.1541882, 10.1145/3312739,
10.1016/j.knosys.2021.106878} and the references therein.

\medskip
Burden in data preparation has been a barrier for companies to 
adopt AI. When it comes to anomaly detection,
unsupervised algorithms require a minimal workload
to prepare and audit the data.
However, this is also associated with the price
that a greater amount of data is needed and
acceptable performance is harder to achieve.

\medskip
In this paper, we consider semi-supervised anomaly
detection (SSAD), where in addition to an unlabeled dataset,
a set of normal samples is available for training.
The problem is described as follows.
Let $\T$ be a set consisting of only normal samples. Also
let $\S$ denote an unlabeled dataset that contains
both normal samples and outliers. It is assumed
that $\S$ contains much more normal samples than outliers.
Given a desired predicted positive (PP) $k$,
the goal is to select $k$ samples from $\S$ in a way to
grasp as many outliers as possible, i.e., to maximize
the true positive rate (TPR) with respect to a fixed $k$.
The choice for PP is application domain dependent. Contributing
factors in determining its value include
scrutiny cost for false positive samples and impact
for neglecting false negative samples.

\medskip
Our approach in solving SSAD is based upon
quadratic multiform separation (QMS), a new concept
recently proposed by Fan et al. \citep{qms}.
QMS is a mathematical model for classification in which one seeks pairwise
separation of the so-called member sets using quadratic
polynomials that satisfy certain cyclic conditions.

\medskip
Evaluation of QMS22 will be hinged on 
the receiver operating characteristic curve, or ROC curve. The ROC curve
is a very popular and powerful tool as a statistical performance
measure in detection/classification theory and hypothesis testing.
It is the graphical plot of true positive rate (TPR)
against false positive rate (FPR).
In addition to analyzing the strength and predictive
power of a classifier, the ROC curve is also capable
of determining optimal threshold settings as
well as making model comparisons through the area under the curve
(AUC).

\medskip
The remainder of the paper is organized as follows:
in Section 2 we give a brief introduction to the concept of QMS,
in Section 3 we provide the design rationale behind which QMS22 was built upon,
in Section 4 we lay out the details of QMS22,
in Section 5 we provide empirical results on the ninety-five datasets,
in Section 6 we make comparisons with serveral top performing
classifiers, and finally, the conclusion is given in Section 7.

\section{Quadratic Multiform Separation}
In this section, we give a brief introduction to
the concept of quadratic multiform separation (QMS).
We refer to \citep{qms} for more background on QMS.

\medskip
The classification problem that QMS aims to solve is defined as follows.
Let $\Omega\subset\R^p$ be a training set containing instances of $x$ from
input-output pair $(x,y)$ generated
by an {\it unknown} membership function $\mu:\Omega\to\M$.
Here $x\in\R^p$ is a vector, $y$ is a class label selected from the set
$\M=\{1,\dots,m\}$ and $m$ is the number of possible classes.
The problem is to find a classifier $\hat{\mu}(\cdot)$, serves as an approximation of
$\mu(\cdot)$, by equating the behavior of $\hat{\mu}(\cdot)$
to the samples in the training set.
For $i\in\M$, let $\Omega_i$ denote the member set defined by
$$\Omega_i=\{x\in\Omega:\mu(x)=i\}$$
A function in the form of
$f(x)=\|Ax-b\|^2$ is called a member function,
where $q$ is a positive integer,
$A\in\R^{q\times p}$, $b\in\R^q$,
and $\|\cdot\|$ denotes the Euclidean norm.
A function is a member function if and only if it
is a (possibly degenerate) second order nonnegative polynomial.

\medskip
QMS consists of $m$ member functions $f_1,\dots,f_m$,
which are constructed based on the training set $\Omega$.
Accordingly, a classifier $\hat{\mu}(\cdot)$ defined by
$$\hat{\mu}(x)\in\{k:f_k(x)\le f_i(x)\text{~for~all~}i\}$$
can be realized. To facilitate the process of
finding suitable member functions, a loss function
that maps the member functions onto a real number must be defined.
The QMS-specific loss function, denoted by $\Phi$, is
\begin{equation}
\Phi=\sum_{i\in\M}\phi_i \label{s2-1}
\end{equation}
where, for $i\in\M$, $\phi_i$ denotes
\begin{equation}
\phi_i=\sum_{x\in\Omega_i}~\sum_{j\in\M,j\ne i}
\max\left\{\alpha,\frac{f_i(x)}{f_j(x)}\right\} \label{s2-2}
\end{equation}
and $\alpha\in[0,1)$.
Finally, the member functions are found via optimization
by using the coordinate perturbation method (CPM).
CPM adjusts the $A_i$'s and $b_i$'s
(as they form the member functions $f_i(x)=\|A_ix-b_i\|^2$)
to reduce the loss function.
The method works as follows. It successively minimizes
along each and every entry of $A_i$'s and $b_i$'s,
one at a time, to find a descending point.
Along each entry, the method considers at most two nearby points,
one along each of two opposite directions. It moves to
a nearby descending point or otherwise does nothing if both
nearby points do not produce a lower value for the loss function.
The method then switches to the next entry and continues.
With careful implementation, CPM can be extremely efficient
and effective. In particular, there is no need to explicitly compute the
loss function $\Phi$ even once during the entire course of optimization.
For each iteration, i.e., scanning all entries once for
all $A_i$'s and $b_i$'s, the computation complexity of CPM
is linear in $p$, $q$, $m$, and the number of samples in $\Omega$
(recall that $p$ is the number
of features and $q$ is the row dimension of $A_i$).

\medskip
In the sequel, a QMS task is referred to the task of finding
$m$ member functions with respect to $m$
member sets by solving the underlying
optimization problem.

\section{Design Rationale}
We consider the task of solving a trivial unsupervised anomaly
detection problem (TUAD) in this section. The purpose here is to provide
our decisions behind which QMS22 was built upon,
and the reasons why those decisions were made.
In this exercise, instead of tackling the problem with a simple-minded approach,
we will solve it differently using
ideas that can be crafted into a plausible scenario for SSAD.

\medskip
Recall that $\S$ is an unlabeled dataset containing
both normal samples and outliers. Assume that the set $\T$ is not present here.
Let $\S_0$ and $\S_1$ respectively represent the set of normal samples
and the set of outliers in $\S$.
The problem TUAD is to find the outliers in $\S$, given the assumption
that $\S_0$ is also available. 
It is clearly visible that the obvious solution
is to check whether a sample lies in $\S$ but not in $\S_0$.

\begin{figure}[!ht]
\centerline{\includegraphics[width=2.5in,height=0.4in]{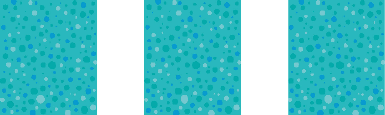}}
\centerline{$\Omega_1=\S_0$\hspace{0.43in}$\Omega_2=\S_0$\hspace{0.43in}$\Omega_3=\S_0$}
\caption{First QMS task. All member sets are
identical and do not contain outliers.}
\label{figure1}
\end{figure}

\medskip
We would like to think TUAD in the context of QMS.
Let us consider a QMS task with three classes and letting
all members sets be equal to $\S_0$, as shown in Figure \ref{figure1}.
Also, we would like to inspect the expression of
the corresponding loss function. To simplify the
discussion, we choose $\alpha=0$ in (\ref{s2-2}). 
It is then easily observed that the loss function,
denoted by $\Phi_1$, is equal to
\begin{equation*}
\Phi_1=\sum_{x\in\S_0}\left\{
\left(\frac{f_1(x)}{f_2(x)}+\frac{f_2(x)}{f_1(x)}\right)
+\left(\frac{f_1(x)}{f_3(x)}+\frac{f_3(x)}{f_1(x)}\right)\right.
\end{equation*}
\begin{equation}
+\left.\left(\frac{f_2(x)}{f_3(x)}+\frac{f_3(x)}{f_2(x)}\right)\right\} \label{x3-1}
\end{equation}
Each parentheses in (\ref{x3-1}) contains
two terms that are reciprocal to each other.
Therefore whenever possible, the optimization for keeping
the loss function low will drive those terms in parentheses
into equal values. That is to say,
at the solution of minimizing $\Phi_1$, it is likely to observe that
\begin{equation}
f_1(x)=f_2(x)=f_3(x)\mbox{ for all  }x\in\S_0 \label{s3-2}
\end{equation}
This result is perhaps expected due to the apparent symmetry of the problem.

\begin{figure}[!hbt]
\centerline{\includegraphics[width=2.5in,height=0.4in]{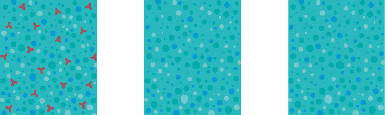}}
\centerline{$\Omega_1=\S$\hspace{0.47in}$\Omega_2=\S_0$\hspace{0.44in}$\Omega_3=\S_0$}
\caption{Second QMS task. Only the first member set contains
outliers.}
\label{figure2}
\end{figure}

\medskip
Now we consider another QMS task as shown in Figure \ref{figure2}.
It is identical to
the first one except $\Omega_1$ is now equal to $\S$.
In this case, the loss function, denoted by $\Phi_2$, becomes
\begin{equation*}
\Phi_2 = \Phi_1 + \sum_{x\in\S_1}\left\{
\frac{f_1(x)}{f_2(x)}+\frac{f_1(x)}{f_3(x)}\right\}
\end{equation*}
The assumption that $\S$ has much more
normal samples than outliers comes to play an important role here.
Consequently, at the solution of minimizing $\Phi_2$, it is likely to observe that
\begin{itemize}
\item $f_1(x)\approx f_2(x)\approx f_3(x)$ for $x\in\S_0$
\item $f_1(x)<f_2(x)$ and $f_1(x)<f_3(x)$ for $x\in\S_1$
\end{itemize}
Therefore, the following function may serve as a good outlier score function:
\begin{equation}
\eta(x)=\left(\frac{f_2(x)-f_1(x)}{f_1(x)}\right)^+
+ \left(\frac{f_3(x)-f_1(x)}{f_1(x)}\right)^+ \label{s4-1}
\end{equation}
where given $a\in\R$, $a^+=a$ if $a>0$ and $a^+=0$ otherwise.
In other words, the magnitude of $\eta(x)$ provides an indication 
on how likely $x$ is an outlier.

\medskip
Now TUAD can be essentially solved as follows:
\begin{enumerate}
\item Carry out the second QMS task and obtain member
functions $f_1$, $f_2$, and $f_3$.
\item For each $x\in\S$, compute its outlier score $\eta(x)$.
\item Take first $n_1$ top outlier score samples as outliers, where
$n_1$ is the cardinality of $\S_1$.
\end{enumerate}

\section{Proposed Method}
How is TUAD related to SSAD that we intend to solve?
First, the set $\S_0$ mentioned in TUAD is of course not available
at our disposal. Nevertheless, the set $\T$
or a subset of $\T$ may serve as an approximation of $\S_0$.
Second, the sets $\Omega_2$ and $\Omega_3$ need not be
identical. Instead, they can be individually an approximation
of $\S_0$.

\medskip
For the QMS task to be involved in QMS22, the
member sets $\Omega_1,\dots,\Omega_m$ will follow rules that
$|\Omega_1|$ is larger than $|\Omega_2|$ and
$|\Omega_2|=\cdots=|\Omega_m|$,
where $|\cdot|$ denotes the cardinality. Under these circumstances,
the corresponding classification problem naturally has imbalanced
member sets to work with. In order to migrate the thought
process given in Section 3 into realism, we will add a weight
to $\phi_1$ to remedy this situation by changing the definition of
loss function $\Phi$ to
\begin{equation*}
\Phi=\frac{|\Omega_2|}{|\Omega_1|}\phi_1 + \sum_{i=2}^m\phi_i
\end{equation*}
Finally, the outlier score for the general $m$ is given by
\begin{equation*}
\eta(x)=\sum_{i=2}^m\left(\frac{f_i(x)-f_1(x)}{f_1(x)}\right)^+ 
\end{equation*}

\medskip
\begin{figure}[!ht]
\centerline{\includegraphics[width=2.2in,height=0.5in]{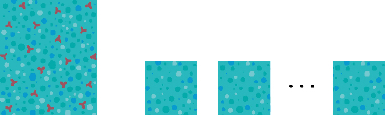}}
\centerline{\hspace{-0.1in}$\Omega_1=\S\cup\T$\hspace{0.26in}
$\Omega_2$\hspace{0.25in}$\Omega_3$\hspace{0.49in}$\Omega_m$}
\caption{The QMS task solved in Algorithm \ref{algo1}.}
\label{figure3}
\end{figure}

\begin{algorithm}[!htb]
\caption{QMS22}
\label{algo1}
\textbf{Inputs}:
\begin{itemize}
\item $\T$: training set, consisting of only normal samples.
\item $\S$: test set, outliers within to be extracted.
\item $\S_0$: subset of $\S$, consisting of normal samples,
used only in plotting the ROC curve.
\item $\S_1$: subset of $\S$, consisting of outliers, $\S=\S_0\cup\S_1$,
used only in plotting the ROC curve.
\item $m$: number of classes for the QMS task.
\end{itemize}
\textbf{Outputs}:
\begin{itemize}
\item ROC curve and its AUC.
\end{itemize}
\begin{algorithmic}[1] 
\STATE (main step) Set $\Omega_1=\S\cup\T$. Shuffle $\T$ well and divide it into
$m-1$ equal parts. Denote them as
$\V_2,\dots,\V_m$, respectively.
For $i=2,\dots,m$, set $\Omega_i=\T-\V_i$.
Solve the QMS task with respect to $\Omega_1,\dots,\Omega_m$
as shown in Figure \ref{figure3}.
Obtain the resulting member functions.
Compute the outlier score for each of the samples in $\S$ using
the function $\eta(\cdot)$ defined in (\ref{s4-1}).
\STATE (generate the ROC curve) For any $\beta\geq 0$, let $\Lambda(\beta)$
be the subset of $\S$ defined by
$$\Lambda(\beta)=\{x\in\S:\eta(x)\geq\beta\}$$
Then the ROC curve is created by plotting
$$\frac{|\Lambda(\beta)\cap\S_1|}{|\S_1|}
\mbox{ versus }
\frac{|\Lambda(\beta)\cap\S_0|}{|\S_0|}$$
at various $\beta$, from a sufficiently large value down to zero.
The algorithm completes.
\end{algorithmic}
\end{algorithm}
We are now ready to present QMS22 as given in Algorithm \ref{algo1}.
Under the framework of QMS,
QMS22 amounts to solve a single $m$-class classification problem
using both the training set $\T$ and the test set $\S$,
where $m$ is a small number (e.g., $3\le m\le 10$).
We use the resulting member functions
to calculate an outlier score for each and every sample in $\S$. This will provide an
estimate on the likelihood of any given sample being an outlier.
Then the ROC curve is subsequently created using the outlier scores.
Notice that outliers only appear in the first member set $\Omega_1$.
Also, all member sets are defined in a way
that intentionally includes a significant amount of overlapping normal samples,
which usually results in normal samples having a lower outlier score
and outliers having a higher outlier score
(see the design rationale in Section 3).

\medskip
Finally, we address the computation complexity of QMS22.
In view of the computation complexity of a generic QMS task mentioned
in Section 2, it is not difficult to see that a similar conclusion can be
made here. That is to say, the computation complexity of QMS22
depends linearly on $n$ (the total number of samples in the training
set and the test set), $p$ (the number of features),
$m$ (a hyperparameter, the number of classes), $q$
(a hyperparameter, the row dimension of $A_i$), and the iteration number in CPM.
Typically, a small $q$ (e.g., 5) and a small iteration number (e.g., 5)
would suffice to produce a decent result.

\section{Performance Evaluation: QMS22 Itself}
In \citep{10.1016/j.knosys.2021.106878}, a thorough
comparison was conducted to provide a broad empirical
study with regard to ninety-five benchmark datasets from the
KEEL repository\footnote{http://sci2s.ugr.es/keel/datasets.php.} \citep{keel} and
twenty-nine state-of-the-art SSAD classifiers. 
Those datasets are not related to a specific domain.
Their experiments lead to the conclusion
that, in terms of the AUC performance,
the top four best players are
BRM \citep{Camia2018BaggingRandomMinerAO},
OCKRA \citep{s16101619},
ISOF \citep{https://doi.org/10.48550/arxiv.1609.06676, 4781136}, and
ocSVM \citep{NIPS1999_8725fb77}.
In this paper we will follow a similar experimental setup
and compare QMS22 against those classifiers.

\medskip
The performance evaluation runs on a computer
with specification given in Table \ref{table3}.
\begin{table}[h!]
\centering
\begin{tabular}{lr} \hline \\[-0.12in]
Processor & Intel Core i7-8700K 3.70GHz × 12 \\
Memory & 62.6 GB \\
OS Name & Fedora 35 64-bit \\ \hline
\end{tabular} 
\caption{Specification of the computer where performance
evaluation is performed.}
\label{table3}
\end{table}
Our code is implemented in the C language using double precision arithmetic.
It only invokes a single user thread and does not use any
GPU. The source code will be available publicly in a later time.

\medskip
The ninety-five datasets are illustrated in Table \ref{table1}.
Each dataset contains five related datasets (namely,
5-fold datasets) to perform cross validation.
Each of the 5-fold datasets consists of a training set and a
test set. Their size ratio is about 4:1.
In our SSAD setup, all outliers in the training
set must be removed prior to the classifier.
Therefore we obtain total of 5 AUC's for the 5-fold datasets.
The representative AUC for the dataset
is then the average of those 5 AUC's.

\medskip
Before sending the data to QMS22, every dataset is preprocessed as follows:
\begin{itemize}
\item Categorical data is converted to a numeric form using OneHotEncoder
\footnote{https://scikit-learn.org/stable/modules/generated/sklearn.preprocessing.OneHotEncoder.html.}.
\item Each feature variable is normalized so that
its maximal absolute value among all samples in
the training set is 255.  
\end{itemize}
The hyperparameters used in QMS22 are listed in Table \ref{table6}.
Finally, The AUC and CPU time results are summarized in
Table \ref{table1}.
\begin{table}[bht]
\centering
\begin{tabular}{ll} \hline \\[-0.12in]
Description & Value \\[0.02in]
\hline \\[-0.1in]
number of classes ($m$)                & 7                                       \\[0.02in]
row dimension of $A_i$ ($q$)           & 10                                      \\[0.02in]
threshold in loss function ($\alpha$)  & 0.5                                     \\[0.02in]
CPM: number of iterations executed     & 60                                      \\[0.02in]
CPM: distance to nearby entry in $A_i$ & 1                                       \\[0.02in]
CPM: distance to nearby entry in $b_i$ & 255                                     \\[0.02in]
CPM: initial condition of $A_i$        & zero matrix                             \\[0.02in]
CPM: initial condition of $b_i$        & 25500 at first entry and zero elsewhere \\[0.02in]
\hline
\end{tabular} 
\caption{Hyperparameters selected in QMS22 (within the QMS task).}
\label{table6}
\end{table}

\section{Performance Evaluation: Comparison}
In this section we compare QMS22 against four above-mentioned
classifiers. The numerical result pertaining to those classifiers
is either taken directly from \citep{10.1016/j.knosys.2021.106878} or borrowed
from the authors of \citep{10.1016/j.knosys.2021.106878}, courtesy of M.E. Villa-P\'{e}rez.
To put the comparison in perspective, we will focus on boxplot
\citep{doi:10.1080/00031305.1978.10479236}, average AUC,
Wilcoxon signed-rank test \citep{doi:https://doi.org/10.1002/9780471462422.eoct979,
pagano2018principles},
and computation complexity. 

\subsection{Boxplots}
\begin{figure}[!ht]
\centering
\includegraphics[]{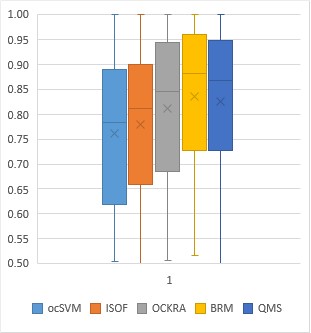}
\caption{Boxplots using all datasets.} 
\label{figure4}
\end{figure}
A boxplot is a method for graphically demonstrating the locality, spread and
skewness groups of numerical data through their quartiles.
It displays the distribution of the data based on a summary of five values:
minimum, first quartile, median, third quartile, and maximum.

\medskip
Based on the plots given in Figure \ref{figure4}, we see that:
\begin{itemize}
\item BRM slightly outperforms QMS22.
\item QMS22 slightly outperforms OCKRA.
\item QMS22 outperforms ISOF and ocSVM.
\end{itemize}

\subsection{Average AUC and standard deviation}
\begin{table}[!ht]
\centering
\begin{tabular}{lllll} \hline \\[-0.12in]
Classifier & Average AUC & STD \\[0.02in]
\hline \\[-0.1in]
QMS22 & 0.8252 & 0.1483 \\
BRM   & 0.8347 & 0.1442 \\
OCKRA & 0.8110 & 0.1491 \\
ISOF  & 0.7785 & 0.1410 \\
ocSVM & 0.7616 & 0.1493 \\[0.02in] \hline
\end{tabular} 
\caption{
Average AUC and the standard deviation (STD) using all datasets.}
\label{table4}
\end{table}
Based on the average AUC given in Table \ref{table4}, it leads to the same
observations as before.

\subsection{Wilcoxon signed-rank tests}
\begin{table}[!ht]
\centering
\begin{tabular}{lllll} \hline \\[-0.12in]
Comparison & $R^+$ & $R^-$ & Null-hypothesis & $p$-value \\[0.02in]
\hline \\[-0.1in]
QMS22 vs. BRM   & 2086 & 2379 & Not rejected & $>0.5$     \\[0.02in]
QMS22 vs. OCKRA & 2446 & 2019 & Not rejected & $>0.4$     \\[0.02in]
QMS22 vs. ISOF  & 3470 & 1090 & Rejected     & $<0.00001$ \\[0.02in]
QMS22 vs. ocSVM & 3481 &  984 & Rejected     & $<0.00001$ \\[0.02in] \hline
\end{tabular} 
\caption{
Wilcoxon signed-ranks tests using all datasets.}
\label{table2}
\end{table}
The Wilcoxon signed-rank tests are conducted to verify the
null-hypothesis that the performances of each paired classifiers,
on all datasets, are similar statistically.
Table \ref{table2} shows the comparisons of
QMS22 against BRM, OCKRA, ISOF,
and ocSVM, respectively. Here $R^+$ is the sum of ranks
for the datasets on which QMS22 outperforms
the paired classifier, and $R^-$ the sum of ranks for the opposite.
With the chosen significance level 0.05, the result of each 
test and $p$-value are stated.  

\medskip
Based on the test results given in Table
\ref{table2} and everything above, we conclude that:
\begin{itemize}
\item There is no statistically significant difference between QMS22 and BRM,
and between QMS22 and OCKRA.
\item QMS22 significantly outperforms ISOF and ocSVM.
\end{itemize}

\subsection{Computation complexities}
The computation complexities for various classifiers are summarized in Table \ref{table5}.
\begin{table}[!ht]
\centering
\begin{tabular}{ll} \hline \\[-0.12in]
Classifier & Complexity \\[0.02in]
\hline \\[-0.1in]
QMS22 & $O(n)$ \\
BRM   & $O(n^2)$ \\
OCKRA & Not reported \\
ISOF  & $O(n)$ \\
ocSVM & Convex quadratic programming \\[0.02in] \hline
\end{tabular} 
\caption{Computation complexities
reported by authors in their work.}
\label{table5}
\end{table}
\begin{table*}[!htb]
\centering
\resizebox{0.99\columnwidth}{!}{
\begin{tabular}{lrrrrrlrrrr} \hline \\[-0.11in]
Dataset & $n$ & $p$ & AUC & $t$ & & Dataset & $n$ & $p$ & AUC & $t$ \\[0.005in]
\hline \\[-0.1in] 
abalone-17\_vs\_7-8-9-10         & 2338 &  8 & 0.7774 & 11.0 & & 
kr-vs-k-zero-one\_vs\_draw       & 2901 &  6 & 0.9711 & 32.1 \\
abalone19                        & 4174 &  8 & 0.6066 & 20.3 & &
kr-vs-k-zero\_vs\_eight          & 1460 &  6 & 0.9350 & 15.7 \\
abalone-19\_vs\_10-11-12-13      & 1622 &  8 & 0.5183 &  7.4 & & 
kr-vs-k-zero\_vs\_fifteen        & 2193 &  6 & 0.9974 & 24.6 \\
abalone-20\_vs\_8-9-10           & 1916 &  8 & 0.7888 &  8.9 & & 
led7digit-0-2-4-5-6-7-8-9\_vs\_1 &  443 &  7 & 0.8877 &  1.4\\
abalone-21\_vs\_8                &  581 &  8 & 0.9543 &  2.5 & & 
lymphography-normal-fibrosis     &  148 & 18 & 0.9717 & 22.5 \\
abalone-3\_vs\_11                &  502 &  8 & 0.9823 &  2.2 & & 
new-thyroid1                     &  215 &  5 & 0.9794 &  0.4 \\
abalone9-18                      &  731 &  8 & 0.8588 &  3.1 & & 
new-thyroid2                     &  215 &  5 & 0.9714 &  0.4 \\
car-good                         & 1728 &  6 & 0.6877 & 12.2 & & 
page-blocks0                     & 5472 & 10 & 0.9478 & 22.9 \\
car-vgood                        & 1728 &  6 & 0.8204 & 12.3 & &
page-blocks-1-3\_vs\_4           &  472 & 10 & 0.9009 &  1.9 \\
cleveland-0\_vs\_4               &  177 & 13 & 0.8187 &  0.9 & & 
pima                             &  768 &  8 & 0.7780 &  1.8 \\
dermatology-6                    &  358 & 34 & 0.9904 &  4.8 & & 
poker-8-9\_vs\_5                 & 2075 & 10 & 0.7376 & 10.1 \\
ecoli-0-1-3-7\_vs\_2-6           &  281 &  7 & 0.8453 &  0.9 & & 
poker-8-9\_vs\_6                 & 1485 & 10 & 0.5851 &  6.9 \\
ecoli-0-1-4-6\_vs\_5             &  280 &  6 & 0.8240 &  0.7 & & 
poker-8\_vs\_6                   & 1477 & 10 & 0.6110 &  6.9 \\
ecoli-0-1-4-7\_vs\_2-3-5-6       &  336 &  7 & 0.8736 &  1.0 & & 
poker-9\_vs\_7                   &  244 & 10 & 0.9150 &  1.1 \\
ecoli-0-1-4-7\_vs\_5-6           &  332 &  6 & 0.9337 &  0.9 & & 
segment0                         & 2308 & 19 & 0.9931 & 16.7 \\
ecoli-0-1\_vs\_2-3-5             &  244 &  7 & 0.9409 &  0.7 & & 
shuttle-2\_vs\_5                 & 3316 &  9 & 0.9986 & 13.9 \\
ecoli-0-1\_vs\_5                 &  240 &  6 & 0.9307 &  0.6 & & 
shuttle-6\_vs\_2-3               &  230 &  9 & 0.9955 &  0.9 \\
ecoli-0-2-3-4\_vs\_5             &  202 &  7 & 0.9359 &  0.6 & & 
shuttle-c0\_vs\_c4               & 1829 &  9 & 0.9987 &  6.7 \\
ecoli-0-2-6-7\_vs\_3-5           &  224 &  7 & 0.9181 &  0.6 & & 
shuttle-c2\_vs\_c4               &  129 &  9 & 0.9757 &  0.6 \\
ecoli-0-3-4-6\_vs\_5             &  205 &  7 & 0.9432 &  0.6 & & 
vehicle0                         &  846 & 18 & 0.9166 &  5.3 \\
ecoli-0-3-4-7\_vs\_5-6           &  257 &  7 & 0.9311 &  0.7 & & 
vehicle1                         &  846 & 18 & 0.7296 &  5.1 \\
ecoli-0-3-4\_vs\_5               &  200 &  7 & 0.9472 &  0.6 & & 
vehicle2                         &  846 & 18 & 0.9437 &  5.2 \\
ecoli-0-4-6\_vs\_5               &  203 &  6 & 0.9509 &  0.5 & & 
vehicle3                         &  846 & 18 & 0.6879 &  5.2 \\
ecoli-0-6-7\_vs\_3-5             &  222 &  7 & 0.9220 &  0.7 & & 
vowel0                           &  988 & 13 & 0.9834 &  5.1 \\
ecoli-0-6-7\_vs\_5               &  220 &  6 & 0.9225 &  0.6 & & 
winequality-red-3\_vs\_5         &  691 & 11 & 0.7959 &  3.3 \\
ecoli-0\_vs\_1                   &  220 &  7 & 0.9821 &  0.3 & & 
winequality-red-4                & 1599 & 11 & 0.5934 &  7.7 \\
ecoli1                           &  336 &  7 & 0.9020 &  0.8 & & 
winequality-red-8\_vs\_6         &  656 & 11 & 0.6258 &  3.1 \\
ecoli2                           &  336 &  7 & 0.8685 &  0.9 & & 
winequality-red-8\_vs\_6-7       &  855 & 11 & 0.5941 &  4.0 \\
ecoli3                           &  336 &  7 & 0.8437 &  1.0 & & 
winequality-white-3-9\_vs\_5     & 1482 & 11 & 0.7271 &  7.4 \\
ecoli4                           &  336 &  7 & 0.9549 &  1.0 & & 
winequality-white-3\_vs\_7       &  900 & 11 & 0.8293 &  4.4 \\
flare-F                          & 1066 & 11 & 0.7825 &  8.3 & & 
winequality-white-9\_vs\_4       &  168 & 11 & 0.5000 &  0.8 \\
glass0                           &  214 &  9 & 0.5578 &  0.6 & & 
wisconsin                        &  683 &  9 & 0.9919 &  2.5 \\
glass-0-1-2-3\_vs\_4-5-6         &  214 &  9 & 0.9581 &  0.7 & & 
yeast-0-2-5-6\_vs\_3-7-8-9       & 1004 &  8 & 0.8054 &  3.3 \\
glass-0-1-4-6\_vs\_2             &  205 &  9 & 0.5000 &  0.7 & & 
yeast-0-2-5-7-9\_vs\_3-6-8       & 1004 &  8 & 0.9068 &  3.3 \\
glass-0-1-5\_vs\_2               &  172 &  9 & 0.5586 &  0.7 & & 
yeast-0-3-5-9\_vs\_7-8           &  506 &  8 & 0.6578 &  1.7 \\
glass-0-1-6\_vs\_2               &  192 &  9 & 0.5000 &  0.7 & & 
yeast-0-5-6-7-9\_vs\_4           &  528 &  8 & 0.7652 &  1.8 \\
glass-0-1-6\_vs\_5               &  184 &  9 & 0.8114 &  0.7 & & 
yeast1                           & 1484 &  8 & 0.7189 &  4.0 \\
glass-0-4\_vs\_5                 &   92 &  9 & 0.8897 &  0.4 & & 
yeast-1-2-8-9\_vs\_7             &  947 &  8 & 0.5817 &  3.3 \\
glass-0-6\_vs\_5                 &  108 &  9 & 0.8234 &  0.4 & & 
yeast-1-4-5-8\_vs\_7             &  693 &  8 & 0.6238 &  2.4 \\
glass1                           &  214 &  9 & 0.6126 &  0.6 & & 
yeast-1\_vs\_7                   &  459 &  7 & 0.7077 &  1.4 \\
glass2                           &  214 &  9 & 0.5000 &  0.8 & & 
yeast-2\_vs\_4                   &  514 &  8 & 0.8918 &  1.7 \\
glass4                           &  214 &  9 & 0.8557 &  0.8 & & 
yeast-2\_vs\_8                   &  482 &  8 & 0.8378 &  1.7 \\
glass5                           &  214 &  9 & 0.6854 &  0.8 & & 
yeast3                           & 1484 &  8 & 0.8671 &  5.0 \\
glass6                           &  214 &  9 & 0.9505 &  0.8 & & 
yeast4                           & 1484 &  8 & 0.7327 &  5.4 \\
haberman                         &  306 &  3 & 0.6311 &  0.4 & & 
yeast5                           & 1484 &  8 & 0.9169 &  5.4 \\
iris0                            &  150 &  4 & 1.0000 &  0.2 & & 
yeast6                           & 1484 &  8 & 0.7744 &  5.5 \\
kr-vs-k-one\_vs\_fifteen         & 2244 &  6 & 0.9999 & 24.5 & & 
zoo-3                            &  101 & 16 & 0.8026 &  0.7 \\
kr-vs-k-three\_vs\_eleven        & 2935 &  6 & 0.9386 & 32.8 & & & & & & \\
\hline
\end{tabular} }
\caption{Datasets.
$n$, $p$, AUC, and $t$ respectively stand for number
of samples, number of features, AUC, and cpu time in second.}
\label{table1}
\end{table*}
\section{Conclusion}
In this paper we propose a simple, of low computation complexity, and yet effective
classifier QMS22 for semi-supervised anomaly detection.
We empirically evaluate its performance on ninety-five imbalanced benchmark
datasets taken from the KEEL repository and compare against
the classifiers BRM, OCKRA, ISOF, and ocSVM, which are the  
top four performing classifiers on the same collection of datasets.
Our study shows that QMS22 significantly outperforms ISOF and ocSVM. Furthermore,
according to the Wilcoxon signed-rank tests,
QMS22 exhibits no statistically significant difference to BRM and OCKRA, respectively.

\section*{Acknowledgements}
The authors wish to thank M.A. Medina-P\'{e}rez for sharing 
the survey paper \citep{10.1016/j.knosys.2021.106878}, and to thank
M.E. Villa-P\'{e}rez for sharing the AUC results
as well as for providing useful clarifications.

\bibliography{Reference}
\end{document}